\pdfoutput=1

\documentclass[11pt]{article}

\usepackage[review]{acl}

\usepackage{times}
\usepackage{latexsym}

\usepackage[T1]{fontenc}

\usepackage[utf8]{inputenc}

\usepackage{microtype}

\usepackage{inconsolata}

\usepackage{graphicx}

\usepackage{times}
\usepackage{soul}
\usepackage{url}
\usepackage{hyperref}
\usepackage[utf8]{inputenc}
\usepackage{caption}
\usepackage{graphicx}
\usepackage{amsmath}
\usepackage{amsthm}
\usepackage{booktabs}
\usepackage{algorithm}
\usepackage{algorithmic}
\usepackage{multicol} 

\usepackage[switch]{lineno}
\usepackage{subcaption}
\usepackage{makecell}
\usepackage{color}
\usepackage{mdframed}
\usepackage{listings}
\usepackage{multirow}
\usepackage{alphalph}

\usepackage{algorithmic}
\newcommand\algorithmicprocedure{\textbf{procedure}}
\newcommand{\algorithmicendprocedure}{\algorithmicend\ \algorithmicprocedure}
\makeatletter
\newcommand\PROCEDURE[3][default]{%
  \ALC@it
  \algorithmicprocedure\ \textsc{#2}(#3)%
  \ALC@com{#1}%
  \begin{ALC@prc}%
}
\newcommand\ENDPROCEDURE{%
  \end{ALC@prc}%
  \ifthenelse{\boolean{ALC@noend}}{}{%
    \ALC@it\algorithmicendprocedure
  }%
}
\newenvironment{ALC@prc}{\begin{ALC@g}}{\end{ALC@g}}
\makeatother

\usepackage{xcolor}
\usepackage{amsfonts}
\usepackage{etoc} 

\DeclareMathOperator*{\argmax}{arg\,max}
\DeclareMathOperator*{\argmin}{arg\,min}

\definecolor{light-gray}{gray}{0.95}

\mdfdefinestyle{prompt}{
  backgroundcolor=light-gray,
  linecolor=light-gray,
  innerleftmargin=1em,
  innertopmargin=0.25em,
  innerbottommargin=0.25em
}

\surroundwithmdframed[style=prompt]{lstlisting}

\title{Zero-Shot Multi-Hop Question Answering\\via Monte-Carlo Tree Search with Large Language Models}

\author{
 \textbf{Seongmin Lee\textsuperscript{1}\thanks{These authors contributed equally to this work}\hspace{1cm}}
 \textbf{Jaewook Shin\textsuperscript{1,3}\footnotemark[1]\hspace{1cm}}
 \textbf{Youngjin Ahn\textsuperscript{2}}
\\
 \textbf{Seokin Seo\textsuperscript{2}}\hspace{1cm}
 \textbf{Ohjoon Kwon\textsuperscript{2}}\hspace{1cm}
 \textbf{Kee-Eung Kim\textsuperscript{1,2}}\vspace{0.12cm}
\\
 \textsuperscript{1}School of Computing, KAIST
 \\
 \textsuperscript{2}Kim Jaechul Graduate School of AI, KAIST
 \\
 \textsuperscript{3}AI Lab, LG Electronics\vspace{0.1cm}
\\
 \texttt{\{tjdals2, jaoc1010, kekim\}@kaist.ac.kr}
\\
}

\begin{document}
\maketitle

\begin{abstract}
    \label{abstract}
    Recent advances in large language models (LLMs) have significantly impacted the domain of multi-hop question answering (MHQA), 
where systems are required to aggregate information and infer answers from disparate pieces of text. 
However, the autoregressive nature of LLMs inherently poses a challenge as errors may accumulate if mistakes are made in the intermediate reasoning steps. 
This paper introduces \textbf{M}onte-Carlo tree search for \textbf{Z}ero-shot multi-hop \textbf{Q}uestion \textbf{A}nswering~(MZQA), a framework based on Monte-Carlo tree search (MCTS) to identify optimal reasoning paths in MHQA tasks, 
mitigating the error propagation from sequential reasoning processes. 
Unlike previous works, we propose a zero-shot prompting method, which relies solely on instructions without the support of hand-crafted few-shot examples that typically require domain expertise. 
We also introduce a behavioral cloning approach (MZQA-BC) trained on self-generated MCTS inference trajectories, 
achieving an over 10-fold increase in reasoning speed with bare compromise in performance.
The efficacy of our method is validated on standard benchmarks such as HotpotQA, 2WikiMultihopQA, and MuSiQue, demonstrating that it outperforms existing frameworks. 


    
\end{abstract}

 \begin{figure*}[ht!]
	\centering
	\includegraphics[width=.95\linewidth]{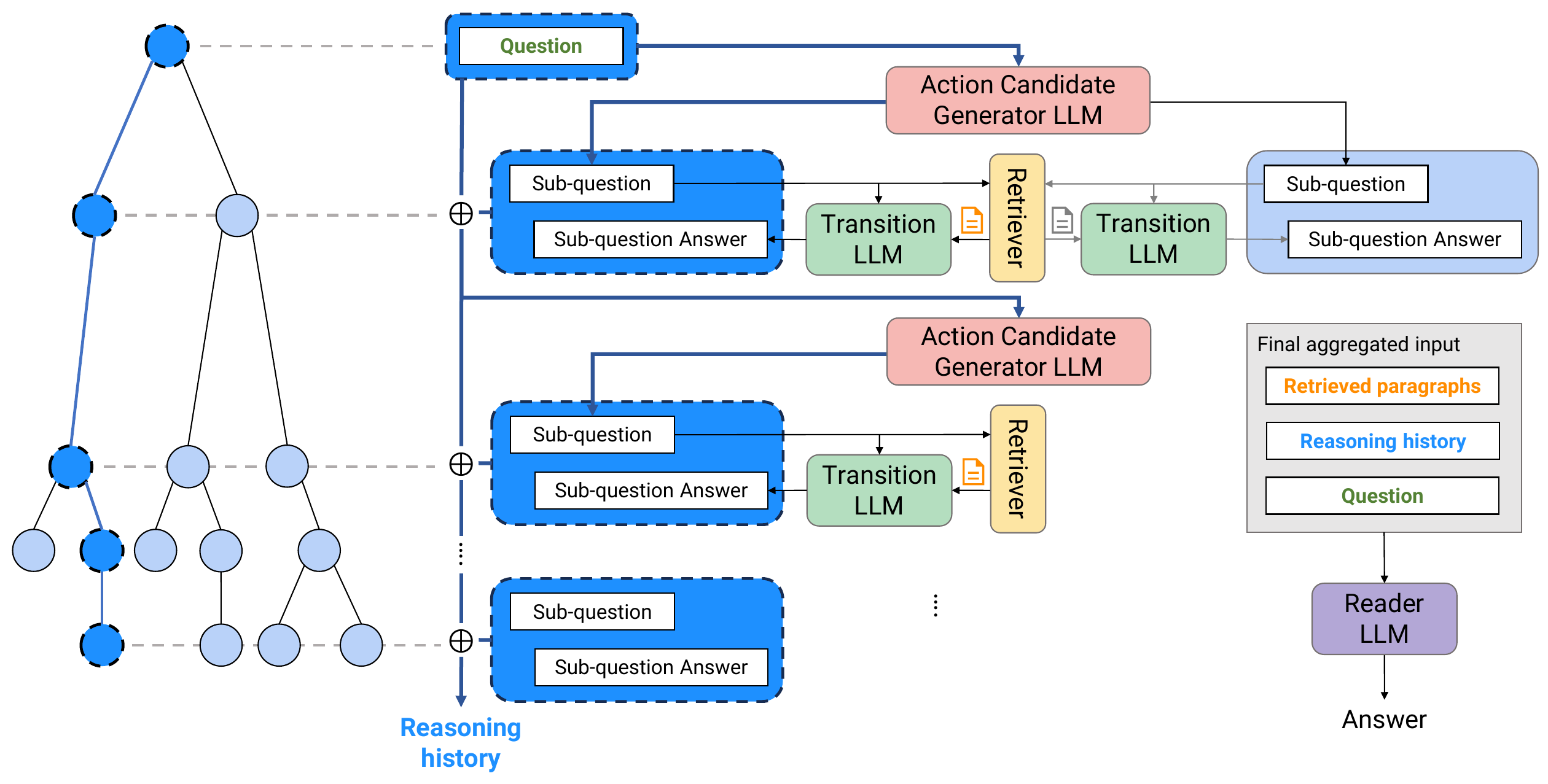}
	\caption{Overview of Monte-Carlo tree search for Zero-shot multi-hop Question Answering (MZQA). }
	\label{fig:framework}
\end{figure*}

\etocdepthtag.toc{mtchapter}
\etocsettagdepth{mtchapter}{none}
\etocsettagdepth{mtappendix}{none}

\section{Introduction}
    \label{sec1}
    Reasoning, a fundamental cognitive process underlying human intelligence, has long been considered a challenging task in the field of artificial intelligence. 
Recent advancements with large language models (LLMs) have shown that not only do models demonstrate strong adaptability and sample efficiency in various natural language tasks~\cite{gpt-3,gpt-4}, 
but also are capable of reasoning when grown to a certain scale and prompted appropriately~\cite{cot,zeroshot-cot}. 
However, many of these showcased tasks can be answered without a complex series of thinking or simply leverage the internal knowledge present in model parameters such as in common sense reasoning or arithmetic.

Multi-hop question answering (MHQA) is a challenging yet critical task in that regard as it requires multiple steps of reasoning through selecting and combining pieces of knowledge to accurately answer a question.
Current approaches to MHQA mostly take the retrieve-and-read paradigm~\cite{retrieve-and-read} that fetches relevant information from external sources on which answers are formulated. 
With LLMs, the chain-of-thought (CoT)~\cite{cot} prompting, which explicitly lays out intermediate reasoning steps, has become a natural choice in dealing with MHQA~\cite{ircot}. 

However, the autoregressive nature of LLMs and CoT poses a significant challenge, as an error at an earlier stage of reasoning can propagate throughout the subsequent steps and adversely affect the outcome.
Tree-structured prompting~\cite{tree-of-thought,rap} has shown promise in mitigating the limitations of LLMs and CoT, but it remains dependent on few-shot examples, making it sensitive to their composition and requiring long input sequences.

In this work, we adopt Monte-Carlo tree search (MCTS) that strikes a balance between exploring potentially valuable thoughts and exploiting already explored thoughts with a zero-shot prompting method.
This allows for circumventing errors at the inception of reasoning and selecting an optimal reasoning path.
However, this approach tends to be compute-intensive given the inherent complexity of MCTS reasoning and requires extensive interactions with LLMs.
To mitigate this, we suggest fine-tuning LLM with behavioral cloning (BC), leveraging the self-generated trajectory of MCTS.
This approach significantly accelerates the reasoning process, all the while minimally compromising the performance.
Moreover, our approach does not necessitate demonstrative examples to the instruction-tuned model, unlike previous approaches.
By doing so, we solely rely on instructional prompts, thereby eliminating the need for hand-crafted examples that may require domain expertise. 

Our contributions are summarized as follows:
(1) We suggest MCTS with zero-shot prompting for MHQA (MZQA), an MCTS-based approach with LLMs for MHQA with a novel zero-shot prompting method that does not require human-generated few-shot examples.
(2) We propose MZQA-BC, a method incorporating BC with reasoning trajectories generated by MCTS, to alleviate computation inefficiencies of MCTS reasoning.
(3) We empirically show that our approaches outperform existing methods in standard MHQA benchmarks, effectively overcoming their limitations. 

\section{Related Work}
    \label{sec2}
    \subsection{LLMs for reasoning} 
Reasoning is a cognitive process that involves drawing conclusions and making decisions based on existing knowledge.
With LLMs, several prompting techniques have been developed to elicit its reasoning capability.
The most notable method is CoT~\cite{cot}, where LLM is prompted with few examples demonstrating a step-by-step reasoning process toward solving a task. 
Its zero-shot variant~\cite{zeroshot-cot} simply provides ``\textit{Let's think step-by-step}'' as a prompt without in-context examples.

While CoT elicits reasoning in LLM, it suffers performance degradation due to its sequential nature; 
if a mistake is made in the intermediate steps, then it will negatively affect the subsequent steps.
One of the ways to circumvent this is self-consistency~\cite{cot_sc}, where the most consistent answer from multiple CoT samples is taken as the answer.
However, this still struggles to solve complex multi-step reasoning tasks where each step is crucial to reasoning correctly.

Recently, tree-based methods such as tree-of-thought (ToT)~\cite{tree-of-thought} and graph-based methods like graph-of-thought (GoT)~\cite{got} have also been proposed.
In these methods, LLM generates and evaluates multiple proposals for the next reasoning step, taking the most promising ones greedily to form the reasoning path.
Reasoning via Planning (RAP)~\cite{rap} and Everything of Thoughts (XoT)~\cite{xot} adopt MCTS for reasoning tasks, similar to our method MZQA. However, MZQA 
distinguishes itself from these two methods in several important aspects. First, they rely on few-shot prompts, in contrast to zero-shot prompts in MZQA, and 
exhibit a significant performance degradation when zero-shot prompts were given. Second, while RAP addresses simple reasoning tasks using only internal knowledge, MZQA is designed for more complex multi-hop QA tasks that require retrieving and integrating information from external knowledge bases. Third, unlike XoT, which utilizes MCTS as an external tool for refining LLM-generated thoughts, we utilizes the LLM inside the MCTS process, generating actions in natural language.
XoT is thus customized for logical reasoning tasks where states and actions can be explicitly defined (i.e. finite sets), making it unsuitable for tasks involving a language action space, e.g. the MHQA task which is the main focus of  our research.




 \begin{figure*}[ht!]
    \vspace{-5pt}
 	\centering
	\includegraphics[width=1\linewidth]{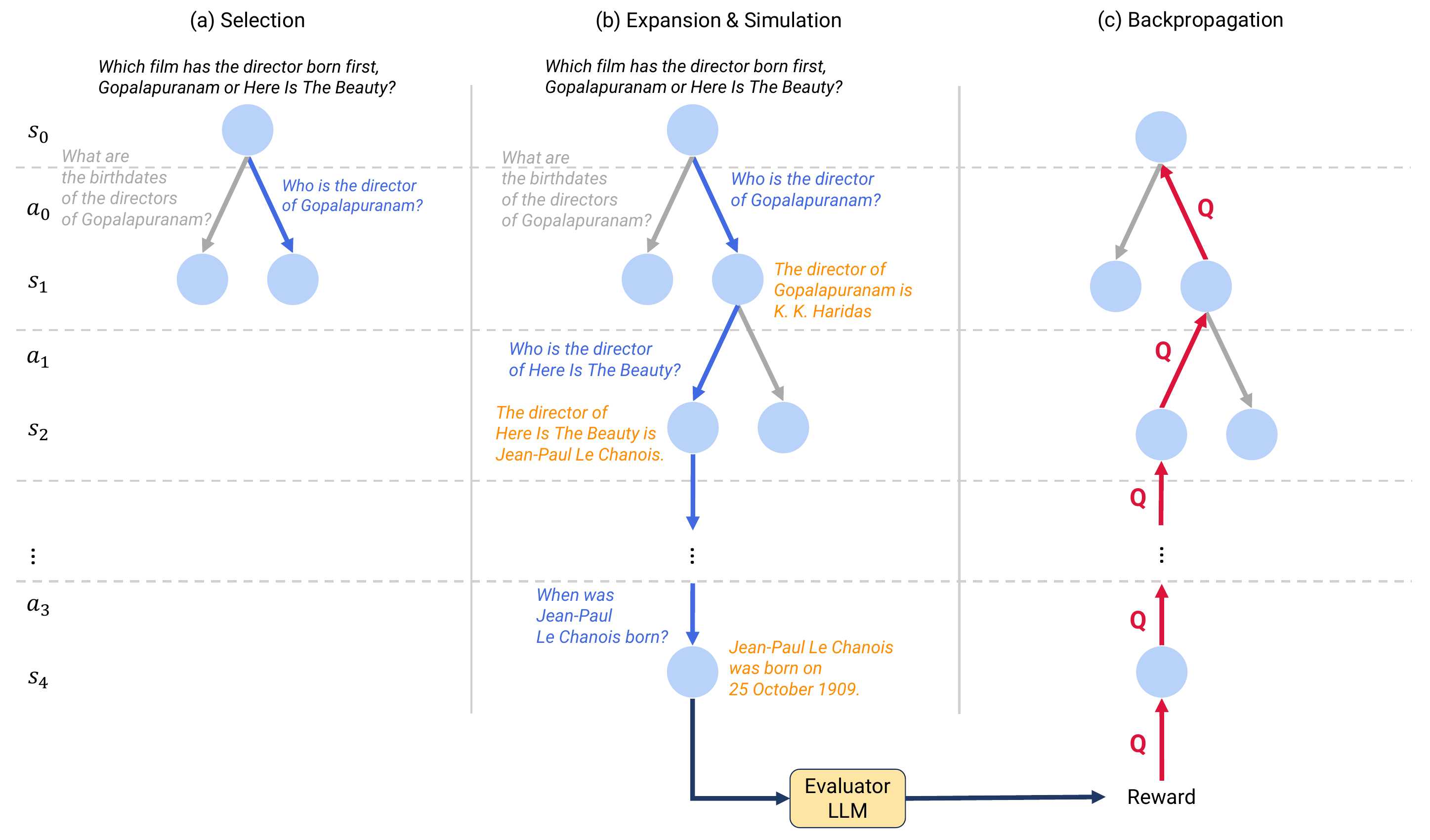}
	\caption{Step-by-step visualization of a single MCTS iteration for multi-hop question answering task.}
	\label{fig:mcts_process}
    \vspace{-5pt}
\end{figure*} 

\subsection{MHQA with LLMs} \label{section:related_work.MHQALLMs}
The current methods for solving MHQA with LLM are dominated by CoT reasoning on the retrieved information.
Interleaving Retrieval CoT (IRCoT)~\cite{ircot} proposes retrieving relevant documents at each step of CoT reasoning instead of retrieving only at a single step, 
helping retrieval with CoT and in turn improving reasoning with fetched documents.
Similarly, ReAct~\cite{react} integrates CoT reasoning and acting via API to enable LLMs to retrieve relevant information when needed.
Multiple Chain Reasoning (MCR)~\cite{MCR} combines multiple IRCoT samples to synthesize a better reasoning history.
Other methods such as DecomP~\cite{decomposed} and self-ask~\cite{self-ask} break down a multi-hop problem into single-hop problems to solve them sequentially.
Due to the aforementioned limitations of CoT-based approaches, SearChain~\cite{SearChain} proposes a DFS tree-based reasoning algorithm designed to reduce errors by evaluating each reasoning step and, if incorrect, reverting to the previous step to generate a new path. However, its greedy approach to pathfinding limits the possibility of finding the optimal path.
LATS~\cite{lats} introduces a method leveraging external knowledge through MCTS. This approach incurs high computational costs, a typical challenge for tree-based searches. It also experiences performance variability, a problem arising from the use of prompting techniques in few-shot contexts, necessitating examples from domain experts.



\section{Preliminaries}
    \label{sec3}
    \subsection{Multi-hop question answering}
Multi-hop question answering (MHQA) aims to answer an intricate goal question, which is difficult to infer directly from a single paragraph.
This problem necessitates multiple steps of accurate reasoning, or \textit{thoughts}, to arrive at a precise answer.
One approach to MHQA is breaking down the goal question into sub-questions, facilitating the retrieval of relevant information from various paragraphs.
Assuming a sufficiently reliable retrieval system, obtaining relevant information through querying sub-questions becomes crucial for accurately deducing the answer. 
Hence, it is important to generate sub-questions that gather useful information missing from current knowledge.  
Our objective is to generate appropriate sub-questions 
that are most helpful in generating the accurate answer to the goal question, given an information retriever.

To illustrate, consider an example with the goal question, ``\textit{Which film has the director born first, Gopalapuranam or Here is the Beauty?}'' (Figure~\ref{fig:mcts_process}). To answer this, the question can be divided into four sub-questions or `hops'. 
The first two hops involve identifying the directors of each movie, followed by determining their birth dates (the next two hops). 
We define a thought in MHQA reasoning as a pair of a sub-question and its answer, for example, Q: ``\textit{Who was the director of Gopalapuranam?}'' and A: ``\textit{The director of Gopalapuranam is K. K. Haridas}''. 
This approach effectively decomposes a complex goal question into manageable sub-questions.

\subsection{MDP formulation of MHQA} 
\label{section:preliminaries.MDP}
We formulate the MHQA problem as an MDP, which is defined as a tuple of $\langle \mathcal{S}, \mathcal{A}, \mathcal{T}, \rho_0, r \rangle$ 
with state space $\mathcal{S}$, action space $\mathcal{A}$, transition model $\mathcal{T}(s_t \mid s_{t-1}, a_{t-1})$, initial state distribution $\rho(s_0)$, reward function $r(s,a,s')$.
\vspace{-3pt}

\paragraph{State}
A state in our scenario is characterized by a sequence of sub-questions and their corresponding answers. 
Formally, an initial state $s_0\in \mathcal{S}$, which consists of a pre-defined prompt and a goal question, is sampled from $\rho_0$.
At timestep $t$, $s_t \in \mathcal{S}$ is a sequence of pairs of questions $a_{i-1}$ and their corresponding answers $z_i$ for $i \in \{1,2,..., t\}$, i.e. $s_t = \{s_0, (a_0, z_1), \ldots, (a_{t-1}, z_t)\}$.
\vspace{-3pt}

\paragraph{Action} 
    Each action $a \in A$ is a sub-question that poses as a query to the retriever.
    To reduce the search space of the large language action space, we exploit LLM as an action candidate generator.

\paragraph{Transition Model} 
    A transition model leverages both the retriever and LLM to generate the next state.
    Specifically, the retriever extracts relevant paragraphs based on the sub-question \(a_{t-1}\).
    Subsequently, LLM generates an appropriate answer \(z_t\) for the sub-question \(a_{t-1}\) with the corresponding retrieved paragraphs and the previous state \(s_{t-1}\). 
    Then, the current state is obtained as $s_t ~ = ~ s_{t-1} ~ \cup ~ \{(a_{t-1}, z_t) \}$. 

    
\paragraph{Reward Function}
    We aim to obtain a successful result without depending on human-generated QA task examples.
    To provide reward signals without these examples, we leverage the internal knowledge of LLM to evaluate the usefulness of the trajectory.
    In particular, at the end of each episode, we query LLM to assess the usefulness of the information acquired by the agent throughout a trajectory in addressing the goal question. 

In this specific MDP, we aim to find an optimal policy $\pi^*: \mathcal{S} \to \Delta \mathcal{A}$ that maximizes cumulative rewards, 
i.e. $\pi^* := \argmax_\pi \mathbb{E_\pi}[\sum_{t=0}^{H-1} r(s_t,a_t,s_{t+1})]$, 
where $\mathbb{E}_\pi$ denotes the expectation given that the agent follows policy $\pi$ and transition model $\mathcal{T}$.

\section{MCTS for Zero-shot MHQA}
    \label{sec4}
    \subsection{MHQA with MCTS}
A major challenge in MHQA is that each hop in the QA process depends on the successful completion of the previous one. 
Errors in the initial reasoning stages often propagate, resulting in incorrect final answers. 
To mitigate this issue, it is imperative to explore potentially beneficial yet previously unexamined reasoning thoughts. 
Therefore, we propose the use of Monte Carlo Tree Search (MCTS) with LLMs based on zero-shot prompting to improve decision-making in MHQA.

With the formulation in Section~\ref{section:preliminaries.MDP}, we employ MCTS for a principled search that exploits options with high value while exploring promising alternatives.
A single iteration of MCTS consists of four main steps~(Figure~\ref{fig:mcts_process}), namely selection, expansion, simulation, and backpropagation. 

\paragraph{Selection} The selection phase finds the most promising sub-question $a_t$ for expansion.
Starting from the root node $s_0$, the algorithm chooses a child node until a leaf node of the current tree is reached.
We employ the commonly used Upper Confidence Bound for Trees (UCT)~\cite{uct}~(Equation \ref{eqn:uct}) as criteria to select the child node, 
which strikes a balance between exploitation of high-value nodes and exploration of less visited nodes. 
If the selected node is terminal, we skip the expansion and proceed directly to backpropagation. 
Denoting $N(s), N(s,a)$ the number of visits to node $s$ and $(s,a)$ respectively, 
$A(s)$ a possible action set in a state $s$,
$\omega$ exploration constant, and $Q(s, a)$ state-action-value of the node, an action is selected by the following rule:
\begin{equation}\label{eqn:uct}
   \text{UCT}(s) = \argmax_{a \in {A}(s) } Q(s,a) + \omega \sqrt{ \frac{ \log N(s) }{ N(s, a) } }
\end{equation}

\paragraph{Expansion} After selecting the node, we expand its children. 
Similar to the proposal prompt in ToT~\cite{tree-of-thought}, we prompt the LLM to generate multiple next sub-questions and retrieve corresponding documents from an external corpus. Then, the transition model outputs the answer to the sub-question using the retrieved paragraphs and the previous state.

\paragraph{Simulation} From each of the expanded nodes, we simulate until the terminal node based on a simulation policy, 
which in our case is generating a single sub-question via LLM to overcome the large language action space.
The reward $r$ is given according to the result of the simulation.

\paragraph{Backpropagation} We update the statistics of the nodes from the expanded node to the root node based on the result of the simulation. 
\vspace{5pt}

After a predetermined number of iterations, the path with the largest total value is chosen as the final reasoning path.\footnote{Although MCTS is typically used as an online planning algorithm, we follow RAP~\cite{rap} for the sake of efficiency.}
Finally, we aggregate the retrieved paragraphs and MCTS reasoning history and 
generate the final answer to the goal question by prompting LLM (called \textit{Reader}) with the final aggregated result.
The overview of the framework is provided in Figure~\ref{fig:framework}.
We refer this as MZQA (\textbf{M}onte-Carlo tree search for \textbf{Z}ero-shot multi-hop \textbf{Q}uestion \textbf{A}nswering).

\subsection{BC with self-generated data}
While MCTS enables the exploration of different reasoning paths, it requires multiple calls to LLM during the search process.
This amounts to requiring more computing resources than left-to-right reasoning methods such as CoT.
In the past literature~\cite{alpagozero,lave},
behavioral cloning (BC), a simple offline imitation learning method that trains the mapping from states to expert actions using given offline expert trajectories, has been considered to mitigate the cost of MCTS inference.
\vspace{5pt}

We apply BC to our setting by fine-tuning LLM with trajectories generated by MCTS regarding as expert trajectories.
Since BC enables the model to mimic the optimal path identified by MCTS without the need for extensive reasoning computations, it can substantially accelerate the decision-making process.

Given initial states $\{s^{(j)}_0\}_{j=1}^m$, $m$ trajectories with a maximum hop $H$ are 
generated by MCTS $ \mathcal{D} := \bigcup_{j=1}^m \{(s^{(j)}_t, a^{(j)}_t)\}_{t=0}^{H-1} $, where $s^{(j)}_t, a^{(j)}_t$ are a state and an action at timestep $t$ of $j$-th trajectory respectively. 
Then, the objective is to minimize the difference between predicted and MCTS actions. 
This can be represented as follows:
\begin{equation} \label{loss}
\pi_{\text{BC}} := \argmin_{\pi} \mathbb{E}_{(s,a) \sim \mathcal{D}} [ L(\pi(s), a) ]
\end{equation}
where $L$ is a supervised learning loss (e.g. cross-entropy loss) between the predicted action $\pi(s)$ and the MCTS action $a$. 
Since the method is a distilled version of MZQA that imitates the MCTS process by tracking the intermediate results,
we call this  MZQA-BC (\textbf{MZQA} with \textbf{B}ehavioral \textbf{C}loning).

\section{Experiments}
    \label{sec5}
    \begin{table*}[ht!]
    \centering
    \vspace{-5pt}
    \resizebox{\textwidth}{!}{
    \begin{tabular}{l cc cccccc cccccccc}
        \toprule[1.5pt]
          & \multicolumn{2}{c}{\textbf{HotpotQA}} & \multicolumn{6}{c}{\textbf{2WikiMHQA}} & \multicolumn{8}{c}{\textbf{MuSiQue}}  \\ \cmidrule(lr){2-3} \cmidrule(lr){4-9} \cmidrule(lr){10-17}
          & \multicolumn{2}{c}{2-hop} & \multicolumn{2}{c}{2-hop} & 
          \multicolumn{2}{c}{4-hop} & \multicolumn{2}{c}{total} & \multicolumn{2}{c}{2-hop} & \multicolumn{2}{c}{3-hop} & \multicolumn{2}{c}{4-hop} & \multicolumn{2}{c}{total} \\ \cmidrule(lr){2-3} \cmidrule(lr){4-5} \cmidrule(lr){6-7} \cmidrule(lr){8-9} \cmidrule(lr){10-11} \cmidrule(lr){12-13} \cmidrule(lr){14-15} \cmidrule(lr){16-17}
         &  F1 & EM & F1 & EM & F1 & EM & F1 & EM & F1 & EM & F1 & EM & F1 & EM & F1 & EM \\ \toprule[1pt]
        IO          & 41.3       &  31.5          &  33.4          &  26.5          &  54.3          &  52.7          &  37.8         &  32.0           & 17.4          & 9.1          & 11.3          & 3.9           & 10.3          & 4.3          & 14.2          & 6.6 \\
        IO (GPT-4) & 48.6          &  35.4         &  41.4          &  32.3          &  66.7          &  64.8          &  46.7          &  39.1           & 26.6          & 15.3          & 23.2          & 16.2           & 20.0          & 10.9          & 24.4          & 14.7 \\

        ToT-BFS    & 40.9  & 31.4  
        & 33.3  & 25.1  & 55.6  & 54.3  & 38.0  & 31.2
        & 25.5  & 15.0  & 16.4  & 8.9  & 9.8  & 3.3  & 19.8  & 10.9\\
        
        ToT-DFS    & 40.0  & 30.1  
        & 30.7  & 23.0  & 52.8  & 51.1  & 35.3  & 28.9
        & 22.5  & 14.2  & 15.6  & 8.9  & 8.0  & 3.3  & 17.7  & 10.6\\
        
        RAP        & 41.0  & 30.7 
        & 35.8  & 26.8  & 55.5   & 54.3   & 39.9  & 32.6 
        & 23.4  & 14.2  & 16.5  & 8.4  & 10.1  & 4.0  & 18.8  & 10.5 \\

        \midrule

        IO w/ Retrieval       & 50.6          &  38.9          &  41.8          &  32.7          &  64.4          &  63.2          &  46.5          &  39.1           & 25.0          & 14.1          & 24.7          & 14.3           & 13.6          & 5.4          & 22.8          & 12.5 \\            
        IO w/ Retrieval (GPT-4)   & 59.5          &  48.1          &  46.8          & 37.5          &  66.8         &  64.5          &  51.0          &  43.1           & 29.2          & 19.2          & 27.6          & 16.9           & \textbf{22.1}         & \textbf{11.3}         & 27.4          & 17.0 \\

        IRCoT          & 54.1          &  42.3          &  55.4          &  42.1          &  72.6          &  70.5          &  59.2          &  48.1           & 37.9          & 26.4          & 23.2          & 14.5           & 16.5          & 6.9          & 29.4          & 19.1 \\
        DecomP         & 37.8          &  29.9          &  52.7          &  43.2          &  73.7          &  73.3          &  57.9          &  50.3           & 19.1          & 10.5          & 9.5           & 4.3            & 15.0          & 10.9         & 15.4          & 8.6  \\

        ToT-BFS w/ Retrieval  & 59.1  & 48.2 
        & 55.4  & 45.2  & 62.2   & 60.3   & 56.8  & 48.4 
        & 40.2  & 30.8  & 27.5  & 18.2  & 14.1  & 6.5  & 31.5  & 22.5 \\

        ToT-DFS w/ Retrieval & 58.1  & 46.6 
        & 54.1  & 45.0  & 60.6   & 58.4   & 55.5  & 47.8 
        & 36.7  & 28.5  & 26.1  & 18.4  & 11.6  & 4.3  & 28.8  & 20.9 \\


        LATS   & 45.6  & 44.9 
        & 54.3  & 53.6  & 69.2   & 69.2   & 57.5  & 56.9 
        & 18.1  & 17.8  & 7.1  & 7.1  & 3.6  & 3.6  & 12.1  & 11.9 \\
         
        MZQA (Ours)    & \textbf{62.3} &  \textbf{49.9} &  67.2          &  56.8          &  71.4          &  70.2          &  68.1          &  59.6           & \textbf{40.8} & \textbf{31.0}          & \textbf{28.9} & \textbf{18.8}  & 17.1 & 8.3 & \textbf{32.8} & \textbf{23.1} \\

        \midrule
        

        MZQA-BC (Ours)   & 59.6  & 47.0 
        & \textbf{67.4}  & \textbf{57.2}  & \textbf{75.1}   & \textbf{73.3}   & \textbf{69.0}  & \textbf{60.6} 
        & 40.8  & 31.0  & 24.4  & 16.9  & 14.2  & 6.1  & 30.9  & 22.1 \\

        \bottomrule[1.5pt]
    \end{tabular}
    }
    \caption{Average F1 and EM (Exact Match) score over 3 runs on MHQA benchmarks by the number of hops in the evaluation samples.
    The top section of the table displays results from baseline models utilizing only the model's internal knowledge, whereas the middle section includes results from baselines that incorporate external knowledge. The bottom section presents the outcomes for the model fine-tuned via BC.
    For standard IO (input-output) prompting, we test both GPT-3.5 (\texttt{gpt-3.5-turbo-1106}) and GPT-4 (\texttt{gpt-4-1106-preview}) with and without retrieval.
    The baselines have been reproduced with GPT-3.5 and the retrieval algorithm used in our method for a fair comparison.}
    \label{tab:odqa_result}
\end{table*}

\begin{figure*}[ht!]
 	\centering
	\includegraphics[width=0.9\linewidth]{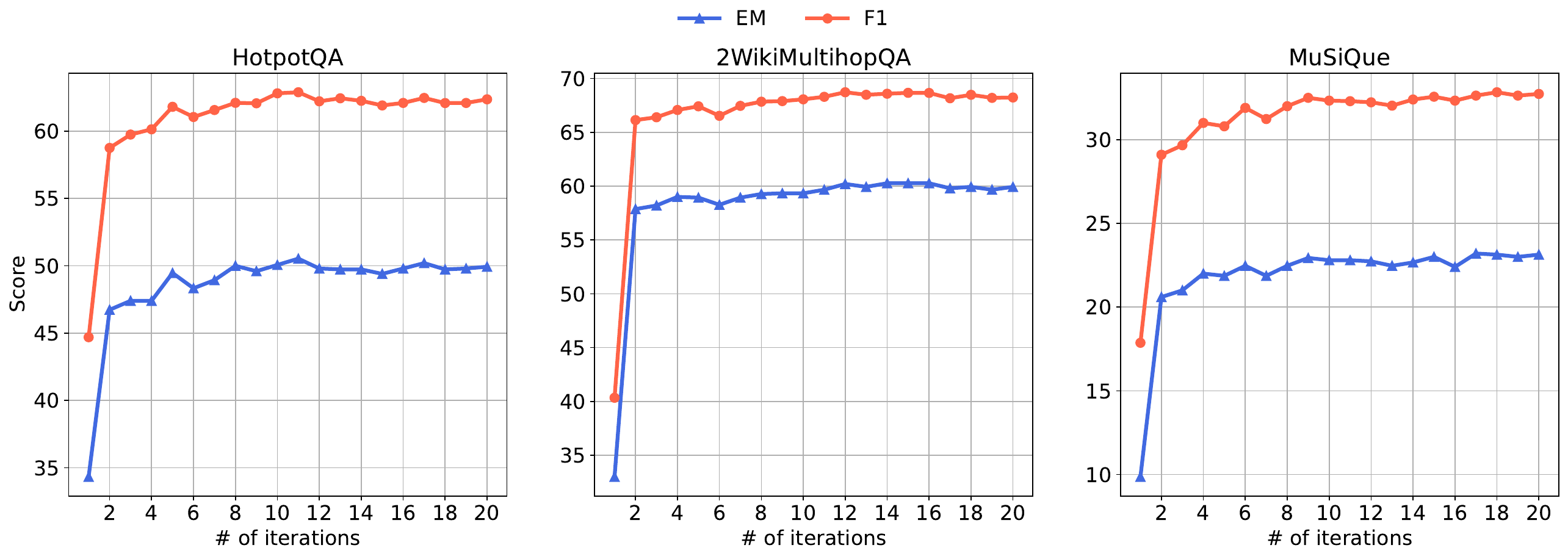}
	\caption{Performance progression of MZQA over iterations averaged across 3 different seeds.}
    \label{fig:mcts_iteration_performance}
\end{figure*}


\subsection{Setup}
\paragraph{Datasets}
The experiments were conducted on three standard open-domain MHQA benchmarks: HotpotQA~\cite{hotpotqa}, 2WikiMultihopQA~\cite{2wiki}, and MuSiQue~\cite{musique}. 
For the hyperparameter search and generating MCTS trajectories, 
we employ 100 data points from the dataset used in IRCoT~\cite{ircot} as a held-out dataset for each benchmark. 
The evaluation involves additional 500 data points from the same dataset for each benchmark.
All benchmarks utilize Wikipedia as the retrieval corpus, consisting of 5,233,329 documents for HotpotQA, 430,225 for 2WikiMultihopQA, and 139,416 for MuSiQue as implemented in IRCoT.



\paragraph{Models}
GPT-3.5~(\texttt{gpt-3.5-turbo-1106})~\cite{gpt-3.5-api} is the default backbone model of all methods unless otherwise stated (\texttt{gpt-4-1106-preview}).
For MCTS, we set the number of iterations to 20 and the episode length (maximum tree depth) to 4, aligning with the maximum hop in our datasets.
In our BC experiments, we used 100 data samples that were excluded from the evaluation set.
We generate trajectories through MCTS by employing these samples, then utilize generated trajectories to fine-tune the model through BC.

\paragraph{Retriever}
We use the retriever based on BM25~\cite{bm25}, implemented using Elasticsearch\footnote{\url{https://www.elastic.co/}} as utilized in the IRCoT~\cite{ircot}.

\paragraph{Baselines}
In our evaluations, we explore three families of established methods to gauge the efficacy of our proposed model in the Multi-Hop Question Answering (MHQA) domain. 
\begin{itemize}
    \item \textbf{Standard IO prompting}: These involve direct question-answering where the model generates responses based solely on the input, and we have tested this with both GPT-3.5 and GPT-4 to ensure alignment with others. 
    \item \textbf{CoT-based methods}: These methods enhance the model's ability to manage complex questions by leveraging intermediate reasoning steps, like IRCoT~\cite{ircot} and DecomP~\cite{decomposed}.  
    \item \textbf{Tree-based methods}: These are hierarchical reasoning approaches such as ToT~\cite{tree-of-thought}, RAP~\cite{rap}, and LATS~\cite{lats}, to provide sophisticated problem-solving capabilities.
\end{itemize}

For IRCoT and DecomP, we use 15 few-shot examples, following the original implementations used to report their experimental results.
For the other few-shot prompting baselines (e.g. ToT, RAP, LATS), we used 5 few-shot examples.
Note that LATS can be seen as an extended version 
of RAP with an additional information retriever, making the experiment a fair comparison to all tree-based baseline
methods with and without retriever settings.
We reproduced all the baselines with GPT-3.5 using 3 seeds. 


\begin{figure*}[t!]
 	\centering
	\includegraphics[width=0.95\linewidth]{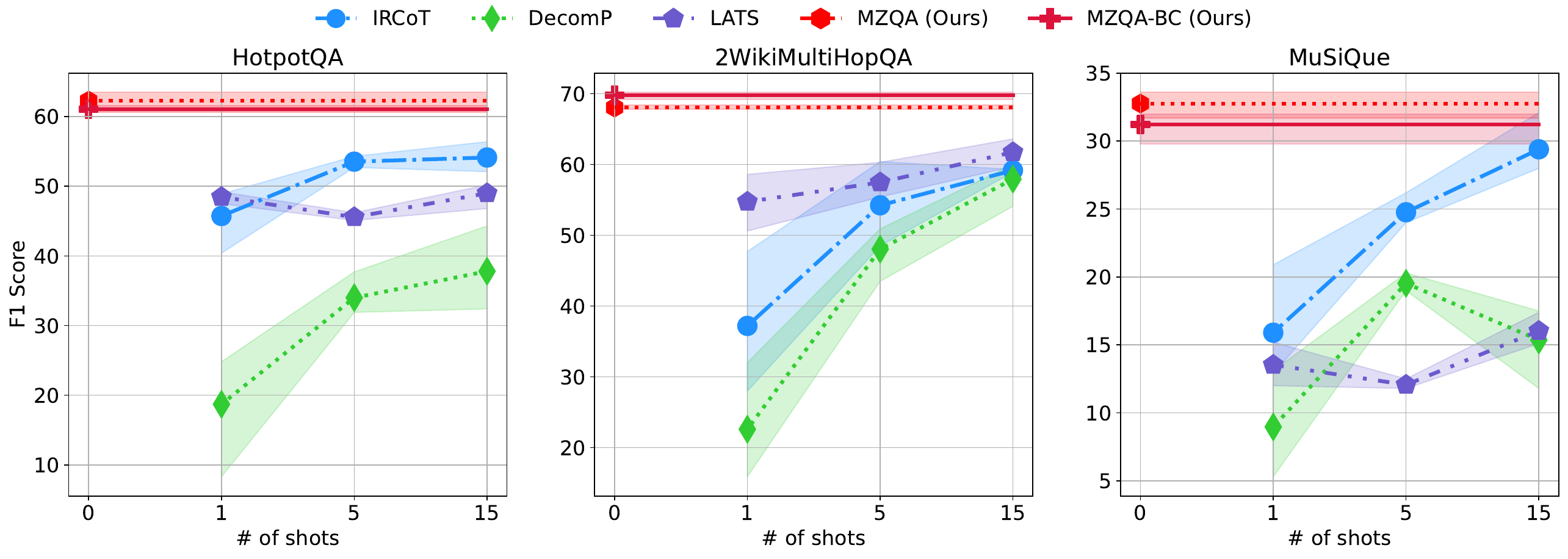}
    \caption{The F1 score with varying numbers of in-context examples (the shaded areas indicate min/max intervals).}
    \label{fig:result_for_few_shot(for f1 score)}
    \vspace{11pt}
\end{figure*}

\begin{figure*}[t!]
 	\centering
	\includegraphics[width=0.95\linewidth]{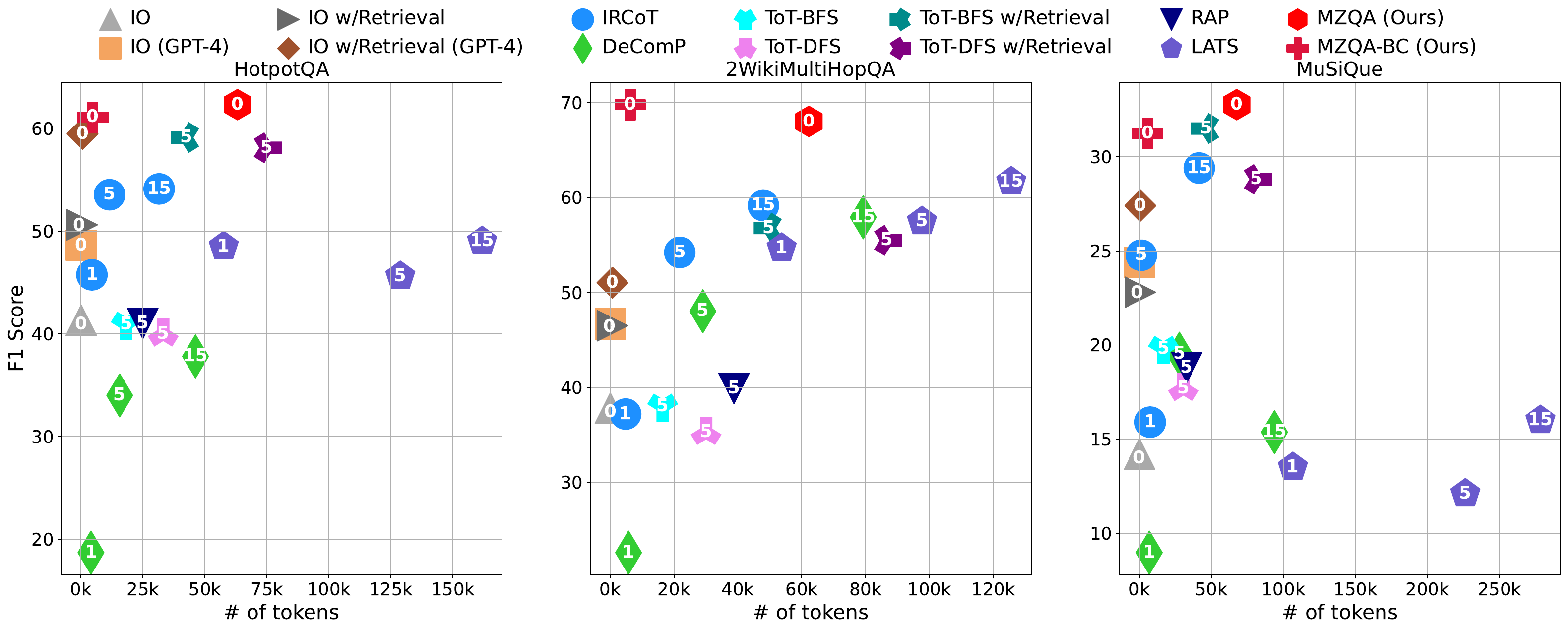}
	\caption{The compute-performance relationship between the average number of tokens required to generate the final answer to the goal question (x-axis) and the average F1 score (y-axis) over 3 seeds on each benchmark. 
 The number written on each data point indicates the number of in-context learning examples.
 Note that the closer the point is to the upper-left corner, the more efficient the method is.}
    \label{fig:tokens}
    \vspace{12pt}
\end{figure*}

\subsection{Performance evaluation}
To evaluate the overall performance of the proposed and baseline methods on MHQA task, 
we measured the statistics of the F1 and exact match (EM) scores on the evaluation dataset from 3 separate runs. 
The results are summarized in Table~\ref{tab:odqa_result}. 

\paragraph{MZQA}
Our MCTS-based reasoning method has demonstrated superior F1 and EM scores compared to the baseline models across all total problem settings in the three MHQA benchmarks. 
Additionally, MZQA approach also outperforms in 7 out of 8 problem settings, with the exception being the 4-hop problem setting in MuSiQue,  
indicating that GPT-4 may contain internal knowledge of the answers corresponding to input goal questions.

\vspace{-1pt}
This result testifies to the effectiveness of our approach in finding optimal reasoning paths through iterative trial and error.
As depicted in Figure~\ref{fig:mcts_iteration_performance}, MZQA improves in the performance with an increase in the number of iterations. 
This indicates that even if the initial iteration does not generate the correct sub-question, the performance is enhanced as MCTS iterates, 
evaluating nodes and selecting better states in the simulation process.

\begin{table*}[t!]
    \centering
    \vspace{-5pt}
    \begin{tabular}{l cc cc cc}
        \toprule[1.5pt]
          & \multicolumn{2}{c}{\textbf{HotpotQA}} & \multicolumn{2}{c}{\textbf{2WikiMHQA}} & \multicolumn{2}{c}{\textbf{MuSiQue}}  \\ \cmidrule(lr){2-3} \cmidrule(lr){4-5} \cmidrule(lr){6-7}
         &  F1 & EM & F1 & EM & F1 & EM \\ \toprule[1pt]

        ToT-BFS (w/ MZQA prompt)   & 60.2 & 48.0  & 64.7 & 56.5  & 32.9 & 23.5 \\
        ToT-DFS (w/ MZQA prompt)  & 60.1 & 47.7  & 61.5 & 53.3  & 31.7 & 21.9 \\
        MCTS (w/ MZQA prompt)    & 62.3 & 49.9  & 68.1 & 59.6  & 32.8 & 23.1 \\
        
        \bottomrule[1.5pt]
    \end{tabular}
    \caption{Performance comparison of different tree-based search algorithms using MZQA prompts across multiple datasets. Each result is averaged over 3 seeds.
    }    
    \label{tab:ablation_study}
\end{table*}

In Figure~\ref{fig:result_for_few_shot(for f1 score)},
we observe that the baseline methods, IRCoT, DecomP and LATS, exhibited significant variance in performance depending on the few-shot prompt, especially when the number of shots is scarce.
In an extreme case, the IRCoT with 1-shot demonstrated a difference between the minimum and maximum F1 score on the 2WikiMultihopQA dataset
is 19.7 points.
In contrast, the maximum difference in the minimum and maximum F1 scores for MZQA  was at most 2.9 points in the worst-case benchmark, HotpotQA. 
This indicates that MZQA shows robust performance with lower variance across multiple runs, 
compared to baselines utilizing few-shot examples.
This is particularly advantageous as it does not rely on few-shot prompts, making our approach more versatile in a wider range of scenarios.

Even though tree search methods (e.g. RAP, ToT, LATS) typically require a significantly larger number of tokens to solve a problem, 
our zero-shot prompting method uses a comparable number of tokens to the CoT-based baseline methods that employ examples (Table~\ref{fig:tokens}). 
In summary, MZQA has demonstrated superior performance in terms of accuracy, consistency, and efficiency, making it a promising approach for tackling MHQA tasks.




\paragraph{MZQA-BC}
MZQA-BC, which is trained on the MCTS-generated trajectories to address computational inefficiencies, 
demonstrated a significant reduction in the number of LLM interactions, leading to faster inference (refer to Table~\ref{tab:visited_state}). 
Furthermore, the number of tokens required to solve a problem was at least ten times less than that required by MZQA, making it comparable to the token usage of the CoT-based baseline methods in a 1-shot setting (refer to Figure~\ref{fig:tokens}). 

The CoT-based baselines may also train a BC model to efficiently generate a more consistent answer, 
but this will require creating a dataset of gold thought trajectories by hand. 

\begin{table}[ht!]
    \centering
    \resizebox{\columnwidth}{!}{\begin{tabular}{lccc}
        \toprule[1.2pt]
          & \textbf{HotpotQA} & \textbf{2WikiMHQA} & \textbf{MuSiQue} \\
        \midrule
        MZQA (Ours)    & 171.9   & 139.4    & 161.5   \\
        MZQA-BC (Ours) & 13      & 13       & 13   \\
        \bottomrule[1.2pt]
    \end{tabular}}
    \caption{The average number of LLM interactions made when answering a goal question. (See Figure~\ref{fig:visualized_mcts} in Appendix for details).} 
    \label{tab:visited_state}    
    \vspace{-5pt}
\end{table}

Interestingly, the performance of MZQA-BC was on par with that of MZQA. 
While there was a slight decrease in the F1 score by 2.7 points for HotpotQA and 1.9 points for MusiQue, an increase of 0.9 points for 2WikiMultihopQA was observed (refer to Table~\ref{tab:odqa_result}). 
This suggests that the BC model can effectively emulate the reasoning process of MZQA, providing a more efficient solution for MHQA tasks without compromising performance.
In contrast to MZQA, although LATS also utilizes MCTS for reasoning with few-shot prompts, the trajectories generated by LATS have empirically proven less effective for fine-tuning LLMs. (See Appendix.~\ref{appendix:BCExperiment})
This highlights the efficacy of MZQA-generated trajectories for LLM fine-tuning. 
In addition, Figure~\ref{fig:result_for_few_shot(for f1 score)} illustrates that MZQA-BC maintains robustness compared to baseline methods, as evidenced by that the difference between the minimum and maximum F1 scores was 2.9 in the worst benchmark, which was HotpotQA.

\subsection{Ablation study}
In this experiment, we aim to evaluate the impact of replacing the search algorithm in MZQA while consistently utilizing the MZQA prompting method.
We assess the efficacy of BFS, DFS, as employed in ToT~\cite{tree-of-thought}, and MCTS with zero-shot prompting across three benchmarks.
As indicated in Table~\ref{tab:ablation_study}, MCTS demonstrated superior performance in HotpotQA and 2WikiMHQA and similar performance in MuSiQue compared to ToT-BFS and ToT-DFS, highlighting its advanced strategic capabilities in structured environments. 
Due to the complex nature of MuSiQue's questions, which pose a significant challenge even to human solvers, all three algorithms exhibited similar effectiveness.

\section{Conclusion}
    \label{sec6}
    We propose MZQA, an approach that combines instruction-based zero-shot prompting with MCTS reasoning to effectively address the challenges of MHQA.
MZQA not only eliminates the need for domain experts to create few-shot examples, 
but also leverages the internal knowledge of LLM with MCTS to exploit the promising reasoning paths while encouraging exploring alternative reasoning paths. 
Furthermore, to enhance the efficiency of path reasoning computation, we introduce MZQA-BC, is a fine-tuning method with MCTS-generated trajectories. 
MZQA-BC mitigates the inefficiencies of MCTS inference, achieving significant improvements in reasoning speed without compromising the quality of the thoughts.


\section*{Limitations}
This study has several key limitations. First, the methods we developed are optimized for large language models, such as GPT-3.5, and may not perform as well with smaller models.
Second, our tree-based approach requires building a new tree for each question. Once a tree is built, it consumes significant computational resources, only to be discarded afterwards. This process can be both costly and inefficient.
Lastly, our fine-tuning process relies heavily on the OpenAI platform, which does not fully disclose how it operates. As a result, we do not have a clear understanding of all the steps involved in how our behavior cloning was fine-tuned.

\section*{Ethical Considerations}
The large language models employed in our study, such as those developed by OpenAI, are prone to generating non-factual outputs due to hallucination phenomena and may also produce biased outcomes, despite the use of advanced techniques like tree search influenced by their pre-training. Furthermore, the reliance on proprietary platforms raises issues of transparency, necessitating a careful approach to the interpretation and dissemination of our findings to thoroughly recognize these constraints. The datasets used in our research, including HotpotQA, 2WikiMultihopQA, and MuSiQue, are licensed under CC BY-SA 4.0\footnote{\url{https://creativecommons.org/licenses/by-sa/4.0/}}, Apache-2.0\footnote{\url{https://www.apache.org/licenses/LICENSE-2.0}}, and CC BY 4.0\footnote{\url{https://creativecommons.org/licenses/by/4.0/}}, respectively.

\bibliography{custom}

\onecolumn
\appendix
\section*{\huge Appendix}
    \label{sec:appendix}
    \makeatletter
\renewcommand*{\thefootnote}{\alph{footnote}}
\makeatother
\setcounter{footnote}{0} 
\setcounter{subsection}{0}

\renewcommand \thepart{}
\renewcommand \partname{}

\etocdepthtag.toc{mtappendix}
\etocsettagdepth{mtchapter}{none}
\etocsettagdepth{mtappendix}{subsubsection}
\setcounter{tocdepth}{3}

\tableofcontents
\newpage
\section{Hyperparameter search}
The performance of open-domain QA can vary by the number of retrieved paragraphs.
Hence, we conduct a hyperparameter search on a validation dataset of 100 samples for each combination of methods, benchmarks, and if applicable the number of in-context learning examples with the options being  $\{2, 4, 6, 8\}$. The optimal number of retrieved documents for each method and benchmark is in Table~\ref{tab:hyperparameter_search}.

For all tree-based baseline models except LATS, the retrieval process utilized the same number of paragraphs as employed by MZQA. The configuration of paragraphs in LATS was designed to adhere to the implementation details specified in the original paper, accurately reflecting its structured prompting methodology.


\begin{table}[h]
    \centering
    \begin{tabular}{lc ccc}
        \toprule[1.5pt]
                                 & \textbf{\# of shots}  & \textbf{HotpotQA} & \textbf{2WikiMHQA} & \textbf{MuSiQue}\\ \midrule[1pt]
        IO w/ Retrieval          & \multirow{2}{*}{-}  & 4   & 6  & 4   \\ 
        IO (GPT-4) w/ Retrieval  &                     & 6   & 6  & 4   \\ \midrule
        \multirow{3}{*}{IRCoT}   & 1                   & 2   & 2  & 4   \\
                                 & 5                   & 2   & 6  & 2   \\
                                 & 15                  & 6   & 4  & 6   \\ \midrule   
        \multirow{3}{*}{DecomP}  & 1                   & 4   & 2  & 6   \\ 
                                 & 5                   & 6   & 4  & 4   \\
                                 & 15                  & 4   & 4  & 2   \\ \midrule
        MZQA                     & \multirow{2}{*}{-}  & 6   & 8  & 6   \\ 
        MZQA-BC                  &                     & 6   & 8  & 6   \\
        \bottomrule[1.5pt]
    \end{tabular}
    \caption{The optimal number of retrieved documents found by hyperparameter search.}
    \label{tab:hyperparameter_search}
\end{table}

\section{Datasets}
\begin{table}[h]
    \centering
    \begin{tabular}{l cccc}
        \toprule[1.5pt]
          & \textbf{2 hop} & \textbf{3 hop} & \textbf{4 hop} & \textbf{corpus size} \\
        \midrule[1pt]
        HotpotQA   & 500   & -    & -    & 5,233,329 \\
        2WikiMHQA  & 395   & -    & 105  & 430,225   \\
        MuSiQue    & 254   & 154  & 92   & 139,416   \\
        \bottomrule[1.5pt]
    \end{tabular}
    \caption{The number of 500 evaluation samples by hops for each dataset on 100 validation samples.}
    \label{tab:dataset_config}
\end{table}

We evaluate the methods on three standard MHQA benchmarks. 
We use the identical 500 evaluation and 100 validation samples as in IRCoT~\cite{ircot}, employing the same retrieval corpora for our experiments.
The breakdown of these evaluation samples is provided in Table~\ref{tab:dataset_config} along with the size of corpora from which documents are drawn.

\section{Implementation details}
The generation parameters are fixed for all methods including the baselines. 
To ensure a more consistent generation, we set the temperature to 0.1 and top-p to 1.0.
Strictly speaking, the transition model in our setting is inherently stochastic, as an answer to a sub-question can take many forms. Nevertheless, we assume that the model is generally deterministic since LLMs usually generate semantically consistent answers.
For MZQA-BC, we use the fine-tuning API provided by OpenAI~\footnote{\url{https://platform.openai.com/docs/guides/fine-tuning}}.
The training hyperparameters are automatically searched by the API.


We adopt several details from the baselines. 
In particular, we use ElasticSearch\footnote{\url{https://www.elastic.co}} with the default search option for BM25-based retriever.
As for the in-context learning examples, 
they are randomly sampled from 20 ready-made demonstrative examples. Hence, the examples may differ from seed to seed.
For tree-based baselines, we standardized the number of action candidates to five, and set the breadth limit of ToT-BFS to one.

\section{Additional BC Experiment}
\subsection{Behavioral cloning applied to other baselines}
\label{appendix:BCExperiment}

\begin{table*}[h]
    \centering
    \vspace{-5pt}
    \begin{tabular}{l c cc cc cc}
        \toprule[1.5pt]
         & \multirow{2}{*}{\textbf{\# of shots}} & \multicolumn{2}{c}{\textbf{HotpotQA}} & \multicolumn{2}{c}{\textbf{2WikiMHQA}} & \multicolumn{2}{c}{\textbf{MuSiQue}}  \\ \cmidrule(lr){3-4} \cmidrule(lr){5-6} \cmidrule(lr){7-8}
         &  &  F1 & EM & F1 & EM & F1 & EM \\ \toprule[1pt]

        IRCoT    & 15  & 54.1 & 42.3 & 59.2 & 48.1 & 29.4 & 19.1 \\
        IRCoT-BC & 15  & 55.3 & 42.5 & 61.1 & 50.3 & 31.3 & 21.7 \\
        \midrule
        LATS  & 15 & 45.6 & 44.9 & 57.5 & 56.9 & 12.1 & 11.9 \\ 
        LATS-BC  & 15  & 29.4 & 29.4 & 41.0 & 41.0 & 5.9 & 5.9 \\ 
        \midrule
        MZQA   & 0  & 62.3 & 49.9 & 68.1 & 59.6 & 32.8 & 23.1 \\
        MZQA-BC   & 0  & 59.6 & 47.0 & 69.0 & 60.6 & 30.9 & 22.1 \\

        \bottomrule[1.5pt]
    \end{tabular}
    \caption{BC performance with trajectories generated by baselines and our method using 3 seeds.}
    \label{tab:additional_bc}
\end{table*}

In this experiment(Table~\ref{tab:additional_bc}), we aim to validate the suitability of trajectories produced by our proposed method and two baseline methods, IRCoT and LATS, for BC.
As anticipated, BC utilizing trajectories generated by IRCoT and MZQA maintains performance levels comparable to those observed prior to fine-tuning.
Conversely, we observe that BC with trajectories generated by LATS results in performance degradation relative to their original trajectory performance.
We hypothesize that this degradation arises because LATS requires strictly structured responses for the given input prompts, thereby limiting the generalization capabilities of LLMs.
These findings suggest that not all prompting methods ensure to generate trajectories that retain BC performance, as evidenced by LATS.
However, our proposed method maintains performance levels similar to the original trajectory performance used for BC, indicating that MZQA-generated trajectories are more suitable for fine-tuning LLMs to address MHQA tasks.

\subsection{Robustness of MZQA-BC on datasets}

\begin{table}[h]
  \centering
  \begin{tabular}{@{}l cccccccc@{}}
    \toprule
     & \multicolumn{2}{c}{\textbf{HotpotQA}} & \multicolumn{2}{c}{\textbf{2WikiMHQA}} & \multicolumn{2}{c}{\textbf{MuSiQue}} \\
     & F1 & EM & F1 & EM & F1 & EM\\
    \midrule
    MZQA  & 62.3 & 49.9 & 68.1 & 59.6 & 32.8 & 23.1 \\
    MZQA-BC w/ HotpotQA & 59.6 & 47.0 & 67.8 & 59.1 & 32.0 & 22.7 \\
    MZQA-BC w/ 2WikiMHQA & 60.3 & 48.0 & 69.0 & 60.6 & 32.8 & 23.2 \\
    MZQA-BC w/ Musique & 59.2 & 47.1 & 67.2 & 58.4 & 30.9 & 22.1 \\
    \bottomrule
  \end{tabular}
  \caption{Comparing BC performance across the datasets, averaged over 3 seeds.}
  \label{tab:additional_bc_2}
\end{table}

To evaluate the generalization capability of the fine-tuned MZQA via BC, We conducted additional experiment(Table~\ref{tab:additional_bc_2}) to asses the performance of MZQA-BC on diffent datasets that were not utilized during BC training. In the table above, the second row shows the performance of MZQA-BC on three datasets, fine-tuned on 100 training instances from the HotpotQA dataset. The results show that MZQA-BC exhibits robust generalization performance over novel datasets.

\newpage

\section{Qualitative Analysis}

\begin{table*}[ht!]    
  
  \centering
    \resizebox{\textwidth}{!}{
  \begin{tabular}{c|c|c|c|c}
    \toprule[1.5pt]
    \textbf{hop} & \textbf{Gold Facts} & \textbf{IRCoT} & \textbf{DecomP} & \textbf{MZQA (Ours)} \\ 
    \midrule[1pt]
    Q & \multicolumn{4}{c}{\makecell[c]{Who is the president of the newly declared independent country part of the Commission\\ of Truth and Friendship with the country that made Arisan!?}}\\
    \midrule
    1 & \makecell[l]{Arisan! is a 2003 \textcolor{blue}{Indonesian} film \\that has drawn more than 100,000 \\viewers.} & \makecell[l]{The president of the newly declared \\independent country part of the \\Commission of Truth and Friendship \\with the country that made Arisan! is\\ \textcolor{blue}{Indonesian} President \textcolor{red}{Susilo Bambang} \\\textcolor{red}{Yudhoyono.}} &\makecell[l]{What is the name of the newly \\declared independent country?\\ \textcolor{red}{Brazil}}  &\makecell[l]{Which country made Arisan?\\\textcolor{blue}{Indonesia}}\\
    \midrule
    2 & \makecell[l]{The Indonesia–\textcolor{blue}{Timor} Leste \\ Commission on Truth and Friendship \\was a truth commission established\\ jointly by the governments of Indonesia \\and \textcolor{blue}{East Timor} in August 2005.} & \makecell[l]{So the answer is: \\\textcolor{red}{Susilo Bambang Yudhoyono.}} &\makecell[l]{Who is the president of Brazil?\\\textcolor{red}{Jair Bolsonaro}}   &\makecell[l]{What is the Commission of Truth \\and Friendship??\\The Commission of Truth and \\Friendship was a truth commission \\established jointly by the \\governments of Indonesia and \\ \textcolor{blue}{East Timor} in August 2005 ...}\\
    \midrule
    3 & \makecell[l]{Democratic Republic of Timor ... \\President \textcolor{blue}{Francisco Guterres} \\Prime Minister Mari Alkatiri ...} & \makecell[c]{-}  &\makecell[c]{-}  &\makecell[l]{What is the name of the newly \\declared independent country?\\The newly declared independent \\country is \textcolor{blue}{East Timor}.}\\
    \midrule
    4 & \makecell[c]{-} & \makecell[c]{-} & \makecell[c]{-}  &\makecell[l]{Who is the president of East Timor?\\The president of East Timor is\\ \textcolor{blue}{Francisco Guterres}}\\
    \bottomrule[1.5pt]
  \end{tabular}
  }
  \caption{
    Comparative analysis of the responses generated by IRCoT, DecomP and MZQA on a MuSiQue dataset example. 
  \textcolor{blue}{Blue} words indicate keywords in generated sentences that match gold facts, while \textcolor{red}{red} words indicate keywords that are irrelevant to gold facts.
  }
  \label{tab:qualitative_analysis}
\end{table*}

To further understand the behavior of our proposed framework, we conduct a qualitative analysis comparing our algorithm with the baseline models. 
We examine the reasoning paths taken by each algorithm to answer a 3-hop question from MuSiQue, 
``\textit{Who is the president of the newly declared independent country part of the Commission of Truth and Friendship with the country that made Arisan!?}"

To answer this example question accurately, the necessary information includes  
(1) the country that made \textit{Arisan!} (\textit{Indonesia}), 
(2) the country that participated in the Commission of Truth and Friendship with \textit{Indonesia} (\textit{East Timor}), and 
(3) the president of \textit{East Timor} (\textit{Franciso Guterres}).
These details are provided in the gold facts column in  Table~\ref{tab:qualitative_analysis}.

The table highlights that IRCoT, 
which bases its retrieval and reasoning on history without generating sub-questions for answering the goal question, tends to prematurely conclude the final answer upon encountering relevant keywords in its historical data. 
Furthermore, DecomP, lacking a mechanism to assess the quality of its generated sub-questions, fails to create relevant sub-questions, leading to incorrect conclusions for the goal question.

In contrast, MZQA produces multiple promising sub-questions using the action candidate generator LLM to reach the correct answer. 
Additionally, by evaluating reasoning paths using the evaluator LLM, 
MZQA can leverage the reward signals to select an optimal path from an extensive set of reasoning paths.
Taking these advantages, MZQA overcomes limitations of CoT-basaed baseline methods and hence finally generates successful a reasoning path, as observed in Table~\ref{tab:qualitative_analysis}.
This underscores the effectiveness of MZQA in handling complex multi-hop reasoning tasks.
(See Figure~\ref{fig:visualized_mcts} for a more detailed reasoning process of MZQA)

\newpage
\section{Pseudocode}

\begin{algorithm}
    \caption{MZQA}
    \label{alg:algorithm}
    \begin{algorithmic}[1]
    \REQUIRE Initial state $s_0$, exploration weight $\omega$, instruction prompt $\mathbf{p}$, action candidates generator $G$, transition model $\mathcal{T}$, reward function $R$, value function $Q$, visitation number of the node $N$
    \begin{multicols}{2}
    \PROCEDURE{Search}{$s_0$}
        \WHILE{within computational budget}
        \STATE $S$ $\xleftarrow{}$ \textsc{Select}($s_0$)
        \FOR{$s \in S$}
        \STATE $r$ $\xleftarrow{}$ \textsc{Simulate}($s$)
        \STATE \textsc{Backpropagation}($s$, $r$)
        \ENDFOR
        \ENDWHILE
        
    \ENDPROCEDURE    
    
    \PROCEDURE{Select}{s}
        \WHILE{$s$ is not a terminal state}
        \IF {$s$ is not expanded}
        \RETURN \textsc{Expand}($s$)
        \ELSE
        \STATE $a \xleftarrow{} \underset{a \in {A}(s) }{\argmax} \space \frac{Q(s,a)}{N(s,a)} + \omega \sqrt{ \frac{ \log N(s) }{ N(s, a) } }$
        \STATE $s' \xleftarrow{} \mathcal{T}(s,a)$
        \STATE $s \xleftarrow{} s'$
        \ENDIF
        \ENDWHILE
    \RETURN $\{s\}$        
    \ENDPROCEDURE

    
    \PROCEDURE{Expand}{$s$}
        \STATE $S := \{ \}$
        \STATE $A \xleftarrow{} G(\mathbf{p}_{\text{expansion}}, s)$ 
        \FOR{$a \in A$}
            \STATE $s^\prime \xleftarrow{} T(s,a)$
            \STATE $S \xleftarrow{} S \cup \{ s^\prime \}$
        \ENDFOR
        \RETURN $S$
    \ENDPROCEDURE
    
    \PROCEDURE{Simulate}{$s$}
        \WHILE{s is non-terminal}
        \STATE $a \sim G(\mathbf{p}_{\text{simulation}}, s)$ 

        \STATE $s^\prime \xleftarrow{} \mathcal{T}(s,a)$
        \STATE $s \xleftarrow{} s'$
        \ENDWHILE
        \RETURN $R(s)$ 
    \ENDPROCEDURE

    \PROCEDURE{Backpropagation}{$s, r$}
        \WHILE{}
        \STATE $N(s) \xleftarrow{} N(s) + 1$
        \STATE $N(s, a) \xleftarrow{} N(s, a) + 1$
        \STATE $Q(s,a) \xleftarrow{} Q(s,a) + r$
        \IF{$s$ is $s_0$}
        \STATE \textbf{break}
        \ENDIF
        \STATE $s \xleftarrow{} \text{parent of } s$
        
        \ENDWHILE
    \ENDPROCEDURE
    \end{multicols}
    \end{algorithmic}  
\end{algorithm}

\newpage
\section{Prompts}
Below are the zero-shot prompts that are used in each component of the proposed framework.

\subsection{Action}
\begin{lstlisting}
Decompose it into the next 'Sub-questions' required to solve the 'Question'.


Consider
- Create "Sub-questions" with keywords from the 'Question' and 'Question history'.


Output format:
Sub-question: answer
\end{lstlisting}
\noindent\textbf{\textit{Example}}\\
\indent\textit{Input}

\begin{lstlisting}
Question: Who is the president of the newly declared independent country part of the Commission of Truth and Friendship with the country that made Arisan!?
\end{lstlisting}

\textit{Output}
\begin{lstlisting}
Sub-question: What is the name of the newly declared independent country?
Sub-question: Who is the president of the newly declared independent country?
Sub-question: What is the Commission of Truth and Friendship?
Sub-question: Which country made Arisan?
\end{lstlisting}

\subsection{Transition model}
\begin{lstlisting}
Answer the 'Sub-question' using the 'Paragraph', 'Evidence', and 'Question' as references.

Consider:
- Sentence: Find the sentence in the 'Paragraph' to answer the 'Sub-question'. If not found, return 'No information'.
- Sub-question Result: Output result for the 'Sub-question'.

Output format as JSON:
{
"Sentence": result,
"Sub-question Result": result
}
\end{lstlisting}

\noindent\textbf{\textit{Example}}\\
\indent\textit{Input}
\begin{lstlisting}
Paragraphs:
Title: Indonesia-Timor Leste Commission of Truth and Friendship
The Indonesia-Timor Leste Commission on Truth and Friendship was a truth commission established jointly by the governments of Indonesia and East Timor in August 2005. The commission was officially created to investigate acts of violence that occurred around the independence referendum held in East Timor in 1999 and sought to find the "conclusive truth" behind the events. ...

.
.
.

Title: Friendship Day
Friendship Day celebrations occur on different dates in different countries. 
\end{lstlisting}
\newpage
\begin{lstlisting}
The first World Friendship Day was proposed for 30 July in 1958, by the World Friendship Crusade. 
On 27 April 2011 the General Assembly of the United Nations declared 30 July as official International Friendship Day. However, some countries, including India, celebrate Friendship Day on the first Sunday of August. In Oberlin, Ohio, Friendship Day is celebrated on 8 April each year.

Evidence:  Which country made Arisan? Indonesia
Question: Who is the president of the newly declared independent country part of the Commission of Truth and Friendship with the country that made Arisan!?
Sub-Question: What is the Commission of Truth and Friendship??
\end{lstlisting}

\textit{Output}
\begin{lstlisting}
{
"Sentence": "The Indonesia-Timor Leste Commission on Truth and Friendship was a truth commission established jointly by the governments of Indonesia and East Timor in August 2005.",
"Sub-question Result": "The Commission of Truth and Friendship was a truth commission established jointly by the governments of Indonesia and East Timor in August 2005 to investigate acts of violence that occurred around the independence referendum held in East Timor in 1999 and sought to find the 'conclusive truth' behind the events."
}
\end{lstlisting}

\subsection{Evaluation}
\begin{lstlisting}
Assign a score from 0 to 10 to indicate how helpful 'Question history' is in solving the 'Question'.

Consider:
- Only assign a score of 10 if the 'Question history' can be utilized to answer the 'Question'.
- Assign a score of 0 if there is no information available to answer the 'Question' using the 'Question history'.

Output format:
Let's think Step by Step: Explain result
Score: answer
\end{lstlisting}
\noindent\textbf{\textit{Example}}\\
\indent\textit{Input}
\begin{lstlisting}
Question: Who is the president of the newly declared independent country part of the Commission of Truth and Friendship with the country that made Arisan!?

Question history:
Sub-question1: Which country made Arisan?
Sub-question Result1: Indonesia
Sub-question2: What is the Commission of Truth and Friendship??
Sub-question Result2: The Commission of Truth and Friendship was a truth commission established jointly by the governments of Indonesia and East Timor in August 2005 to investigate acts of violence that occurred around the independence referendum held in East Timor in 1999 and sought to find the 'conclusive truth' behind the events.
Sub-question3: What is the name of the newly declared independent country?
Sub-question Result3: The newly declared independent country is East Timor.
Sub-question4: Who is the president of East Timor?
Sub-question Result4: The president of East Timor is Francisco Guterres
\end{lstlisting}
\newpage
\textit{Output}
\begin{lstlisting}
Let's think Step by Step: The question history provides all the necessary information to answer the question about the president of the newly declared independent country, East Timor, which is part of the Commission of Truth and Friendship with Indonesia. The president of East Timor is Francisco Guterres.

Score: 10
\end{lstlisting}

\subsection{Reader}
\begin{lstlisting}
Answer 'Question' in words by referring the 'Paragraph' and 'Evidence'.

Consider:
- Keep your answer as short as possible, 10 words or less.
- Don't explain anything, just output the answer.

Output format
Answer: result
\end{lstlisting}

\noindent\textbf{\textit{Example}}\\
\indent\textit{Input}
\begin{lstlisting}
Paragraphs:
Title: Arisan!
Arisan! is a 2003 Indonesian film that has drawn more than 100,000 viewers. It is the first Indonesian film with a gay theme, and the first Indonesian film to use high-definition color enhancement. It uses a mixture of English, standard Indonesian and Jakartan slang. "Arisan!" became the second film in Indonesian film history to win all six major awards in Festival Film Indonesia (FFI), including Best Picture, Best Director, Best Actor, Best Actress, Best Supporting Actor, and Best Supporting Actress, after "Ibunda" in 1986. ...

...

Title: Indonesia-Timor Leste Commission of Truth and Friendship
The Indonesia-Timor Leste Commission on Truth and Friendship was a truth commission established jointly by the governments of Indonesia and East Timor in August 2005. 
The commission was officially created to investigate acts of violence that occurred around the independence referendum held in East Timor in 1999 and sought to find the "conclusive truth" behind the events. ...

Evidence: 
Question: Who is the president of the newly declared independent country part of the Commission of Truth and Friendship with the country that made Arisan!?
Which country made Arisan? Indonesia
...
Who is the president of East Timor? The president of East Timor is Francisco Guterres

Question: Who is the president of the newly declared independent country part of the Commission of Truth and Friendship with the country that made Arisan!?
\end{lstlisting}

\textit{Output}
\begin{lstlisting}
Answer: Francisco Guterres
\end{lstlisting}
\newpage

\subsection{Simulation policy and BC action}
\begin{lstlisting}
Generate one 'Sub-question' to solve the 'Question'.

Consider
- Create a "Sub-question" with keywords from the 'Question' and 'Question history'.

Output format:
Sub-question: answer
\end{lstlisting}
\noindent\textbf{\textit{Example}}\\
\indent\textit{Input}
\begin{lstlisting}
Question: Who is the president of the newly declared independent country part of the Commission of Truth and Friendship with the country that made Arisan!?
\end{lstlisting}

\textit{Output}
\begin{lstlisting}
Sub-question: What is the newly declared independent country?
\end{lstlisting}

\newpage

\section{MCTS visualization}
\begin{figure*}[h!]
 	\centering	\includegraphics[width=1\linewidth]{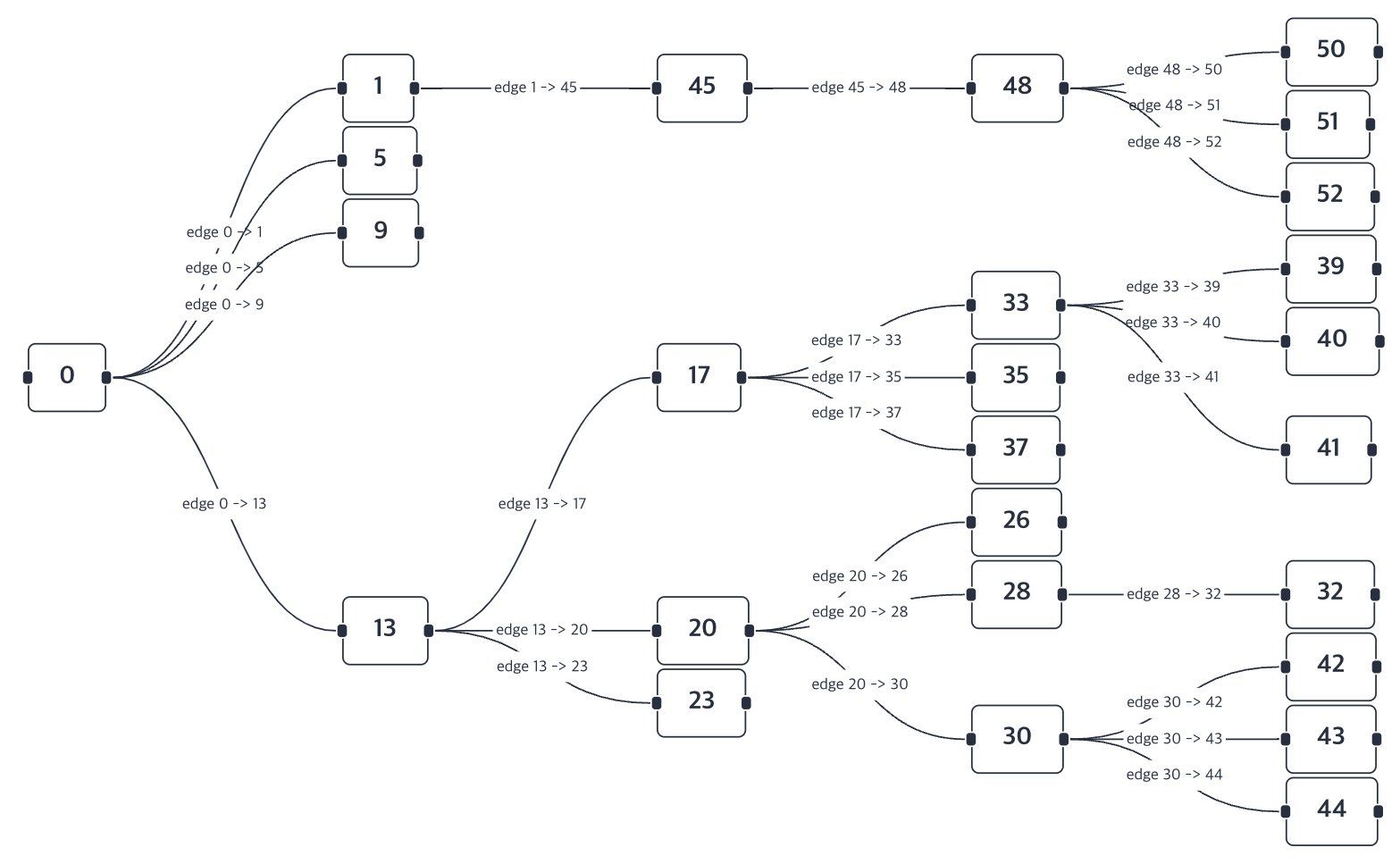}
	\caption{Visualization of the results from 20 iterations of MCTS applied to an instance of the MuSiQue dataset, where the numbers represent sequentially generated nodes. Note that some numbers are removed because those nodes were generated during the simulation process.}
	\label{fig:visualized_mcts}
\end{figure*}

The search tree from a completed MCTS is visualized in Figure~\ref{fig:visualized_mcts}.
For the state associated with each node, we also provide the final sub-question, its corresponding answer, and the reward. 

\paragraph{$\text{Node 0}$}
\begin{lstlisting}
Question: Who is the president of the newly declared independent country part of the Commission of Truth and Friendship with the country that made Arisan!??
\end{lstlisting}

\paragraph{$\text{Node 1}$}
\begin{lstlisting}
Sub-question: What is the name of the newly declared independent country?

Answer: The newly declared independent country is Guinea-Bissau.

Reward: 0
\end{lstlisting}

\paragraph{$\text{Node 5}$}
\begin{lstlisting}
Sub-question: Who is the president of the newly declared independent country?

Answer: The president of the newly declared independent country is not mentioned in the provided paragraph.

Reward: 0
\end{lstlisting}

\newpage
\paragraph{$\text{Node 9}$}
\begin{lstlisting}
Sub-question: What is the Commission of Truth and Friendship?

Answer: The Commission of Truth and Friendship was a truth commission established jointly by the governments of Indonesia and East Timor in August 2005 to investigate acts of violence that occurred around the independence referendum held in East Timor in 1999 and sought to find the 'conclusive truth' behind the events.

Reward: 0
\end{lstlisting}

\paragraph{$\text{Node 13}$}
\begin{lstlisting}
Sub-question: Which country made Arisan?

Answer: Indonesia

Reward: 1
\end{lstlisting}

\paragraph{$\text{Node 17}$}
\begin{lstlisting}
Sub-question: Who is the president of the newly declared independent country?

Answer: The president of the newly declared independent country is not mentioned in the provided paragraph.

Reward: 0.5
\end{lstlisting}

\paragraph{$\text{Node 20}$}
\begin{lstlisting}
Sub-question: What is the Commission of Truth and Friendship?

Answer: The Commission of Truth and Friendship was a truth commission established jointly by the governments of Indonesia and East Timor in August 2005 to investigate acts of violence that occurred around the independence referendum held in East Timor in 1999 and sought to find the 'conclusive truth' behind the events.

Reward: 1
\end{lstlisting}

\paragraph{$\text{Node 23}$}
\begin{lstlisting}
Sub-question: What is the relationship between the newly declared independent country and the country that made Arisan?

Answer: The relationship between the newly declared independent country and the country that made Arisan is not mentioned in the provided text.

Reward: 0
\end{lstlisting}

\paragraph{$\text{Node 26}$}
\begin{lstlisting}
Sub-question: Who is the president of the newly declared independent country?

Answer: The president of the newly declared independent country is not mentioned in the provided text.

Reward: 0
\end{lstlisting}
\newpage
\paragraph{$\text{Node 28}$}
\begin{lstlisting}
Sub-question: What is the name of the newly declared independent country?

Answer: The newly declared independent country is East Timor

Reward: 1
\end{lstlisting}

\paragraph{$\text{Node 30}$}
\begin{lstlisting}
Sub-question: What is the role of the Commission of Truth and Friendship in relation to the newly declared independent country?

Answer: The role of the Commission of Truth and Friendship in relation to the newly declared independent country is to investigate acts of violence that occurred around the independence referendum held in East Timor in 1999 and seek to find the 'conclusive truth' behind the events.

Reward: 1
\end{lstlisting}

\paragraph{$\text{Node 32}$}
\begin{lstlisting}
Sub-question: Who is the president of East Timor?

Answer: The president of East Timor is Francisco Guterres.

Reward: 1
\end{lstlisting}

\paragraph{$\text{Node 33}$}
\begin{lstlisting}
Sub-question: What is the name of the newly declared independent country?

Answer: The name of the newly declared independent country is not mentioned in the provided paragraph.

Reward: 1
\end{lstlisting}

\paragraph{$\text{Node 35}$}
\begin{lstlisting}
Sub-question: What is the Commission of Truth and Friendship?

Answer: The Commission of Truth and Friendship was a truth commission established jointly by the governments of Indonesia and East Timor in August 2005 to investigate acts of violence 
that occurred around the independence 
\end{lstlisting}

\begin{lstlisting}
referendum held in East Timor in 1999 and sought to find the 'conclusive truth' behind the events.

Reward: 0
\end{lstlisting}

\paragraph{$\text{Node 37}$}
\begin{lstlisting}
Sub-question: What is the significance of the country that made Arisan in this context?

Answer: The significance of the country that made Arisan in this context is that it is the first Indonesian film with a gay theme and the first Indonesian film to use high-definition color enhancement.

Reward: 0
\end{lstlisting}
\newpage
\paragraph{$\text{Node 39}$}
\begin{lstlisting}
Sub-question: What is the name of the newly declared independent country?

Answer: The name of the newly declared independent country is not mentioned in the provided paragraph.

Reward: 0
\end{lstlisting}

\paragraph{$\text{Node 40}$}
\begin{lstlisting}
Sub-question: Who is the president of the newly declared independent country?

Answer: The president of the newly declared independent country is not mentioned in the provided paragraph.

Reward: 0
\end{lstlisting}

\paragraph{$\text{Node 41}$}
\begin{lstlisting}
Sub-question: What is the Commission of Truth and Friendship?

Answer: The Commission of Truth and Friendship was a truth commission established jointly by the governments of Indonesia and East Timor in August 2005 to investigate acts of violence that occurred around the independence referendum held in East Timor in 1999 and sought to find the 'conclusive truth' behind the events.

Reward: 0
\end{lstlisting}

\paragraph{$\text{Node 42}$}
\begin{lstlisting}
Sub-question: Who is the president of the newly declared independent country?

Answer: No information

Reward: 0.2
\end{lstlisting}

\paragraph{$\text{Node 43}$}
\begin{lstlisting}
Sub-question: What is the name of the newly declared independent country?

Answer: The newly declared independent country is East Timor.

Reward: 0
\end{lstlisting}

\paragraph{$\text{Node 44}$}
\begin{lstlisting}
Sub-question: What is the relationship between the newly declared independent country and the Commission of Truth and Friendship?

Answer: The relationship between the newly declared independent country and the Commission of Truth and Friendship is that the commission was established to investigate acts of violence that occurred around the independence referendum held in East Timor in 1999 and sought to find the 'conclusive truth' behind the events.

Reward: 0.5
\end{lstlisting}
\newpage
\paragraph{$\text{Node 45}$}
\begin{lstlisting}
Sub-question: Who is the president of Guinea-Bissau?

Answer: No information

Reward: 0.7
\end{lstlisting}

\paragraph{$\text{Node 48}$}
\begin{lstlisting}
Sub-question: Who made the declaration of independence for the newly declared independent country?

Answer: The declaration of independence for the newly declared independent country was made by the prime minister and president of the Democratic Republic of Vietnam (North Vietnam) since 1945.

Reward: 1
\end{lstlisting}

\paragraph{$\text{Node 50}$}
\begin{lstlisting}
Sub-question: What is the name of the Commission of Truth and Friendship with the country that made Arisan?

Answer: The name of the Commission of Truth and Friendship with the country that made Arisan is
the Indonesia-Timor Leste Commission on Truth and Friendship.

Reward: 1
\end{lstlisting}

\paragraph{$\text{Node 51}$}
\begin{lstlisting}
Sub-question: What is the name of the country that made Arisan?

Answer: The country that made Arisan is Indonesia.

Reward: 0.1
\end{lstlisting}

\paragraph{$\text{Node 52}$}
\begin{lstlisting}
Sub-question: Who is the prime minister and president of the Democratic Republic of Vietnam (North Vietnam) since 1945?

Answer: The prime minister and president of the Democratic Republic of Vietnam (North Vietnam) since 1945 is Ho Chi Minh.

Reward: 0
\end{lstlisting}


\end{document}